\documentclass{article}

\PassOptionsToPackage{numbers}{natbib}

\usepackage[preprint]{neurips_2026}

\usepackage[utf8]{inputenc}
\usepackage[T1]{fontenc}
\usepackage{hyperref}
\usepackage{url}
\usepackage{booktabs}
\usepackage{graphicx}
\usepackage{placeins}
\usepackage{amsfonts}
\usepackage{amsmath}
\usepackage{nicefrac}
\usepackage{microtype}
\usepackage{xcolor}


\title{TreeWY: Speculative Verification for Gated DeltaNet Hybrids}

\author{%
  Sneha Murthy Ghantasala \\
  Thomson Reuters \\
  \texttt{sneha.ghantasala@thomsonreuters.com} \\
}

\begin{document}

\hypersetup{
  pdftitle={TreeWY: Speculative Verification for Gated DeltaNet Hybrids},
  pdfauthor={Sneha Murthy Ghantasala}
}

\maketitle

\begin{abstract}
Modern open models are \emph{hybrids}: most layers are
linear-attention (Gated DeltaNet, GDN) layers carrying a small fixed-size recurrent
state instead of a growing key-value (KV) cache. This makes ordinary decoding
memory-efficient, but hurts speculative decoding. To verify a batch of draft
tokens and then roll back the rejected ones, today's systems snapshot the full recurrent
state at \emph{every} draft position for GDN layers, and those snapshots cannot be shared across
branches of a draft tree, so a wide, high-acceptance tree becomes memory-infeasible. We
remove the snapshots. Using a \emph{tree-structured WY transform} of the gated delta
rule, we compute every draft node's output with a single triangular solve and
reconstruct \emph{only} the one accepted state on commit, storing a small pseudo-value
matrix instead of per-node states; the derivation depends only on the gated delta rule,
not on any other architectural detail. In serving benchmarks on two scales of one
hybrid model family (Qwen3.5 35B and 397B) this cuts speculative recurrent-state memory
and KV-cache pressure at essentially identical acceptance length, turning the freed HBM into higher
throughput and much lower time-to-first-token (TTFT) wherever memory binds, and costing a
few percent where it does not. For tree width the same memory buys affordability: a wider, higher-acceptance draft becomes possible, though not
yet a throughput win.
\end{abstract}

\section{Background}

\textbf{Speculative decoding.} Autoregressive decoding emits one token per forward pass, so it is
memory-bandwidth bound and the GPU sits mostly idle. Speculative
decoding~\citep{leviathan2023spec,chen2023specsampling} fills that
idle compute: a cheap drafter (a small model, or a lightweight ``MTP'' head shipped with
the target) proposes $k$ candidate tokens; the target verifies all $k$ in one pass and
accepts the longest prefix it would have generated itself, leaving the output
distribution unchanged but emitting several tokens per pass. A \emph{tree} drafter
proposes multiple alternatives per position~\citep{cai2024medusa}.

\textbf{Hybrid models and GDN layers.} A \emph{hybrid} model mixes softmax-attention with linear-attention layers (for e.g., Gated DeltaNet, GDN). A softmax
layer keeps a KV cache: it stores a key and value vector for \emph{every} previous token,
so its memory grows linearly with context. A \emph{Gated DeltaNet}
(GDN) layer~\citep{yang2025gdn} instead
maintains one recurrent \emph{state} matrix $S \in \mathbb{R}^{d_v \times d_k}$ per head
summarizing the whole prefix in fixed size. For token $t$ with key $k_t$, value $v_t$,
query $q_t$, scalar decay gate $\alpha_t\!\in\!(0,1)$ and write strength
$\beta_t\!\in\!(0,1)$, the state evolves by the \textbf{gated delta rule}
\begin{equation}
S_t \;=\; \alpha_t\, S_{t-1}\bigl(I - \beta_t k_t k_t^\top\bigr) \;+\; \beta_t v_t k_t^\top,
\qquad o_t = S_t\, q_t. \label{eq:gdn}
\end{equation}
The transition $T_t = \alpha_t(I - \beta_t k_t k_t^\top)$ is a \emph{scalar decay}
times a \emph{rank-1 correction}: the $\beta_t k_t k_t^\top$ term ``erases'' what
the state already predicts for $k_t$ before writing the new value (the ``delta''
rule). Crucially, $S$ is a \emph{lossy summary} of the whole prefix. It cannot
be truncated or partially rolled back the way a KV cache can. That single fact is
the source of the speculative-decoding problem.

\section{The memory problem}
\label{sec:memory}

We make the contrast concrete on the two Qwen3.5~\citep{qwen2025tech} hybrids we
evaluate, both $3{:}1$
GDN-to-softmax with $d_k\!=\!d_v\!=\!128$ and softmax $d_{\text{head}}\!=\!256$, 2 KV
heads: \emph{35B-A3B} (30 GDN, 32 value heads; 10 softmax) and \emph{397B-A17B} (45 GDN,
64 value heads; 15 softmax).

\textbf{Normal decoding.} Summed over the softmax layers, the KV cache costs
$2(\text{K,V}) \times n_{\text{softmax}} \times n_{\text{KV-head}} \times
d_{\text{head}} \times 2$~B (bf16) per token: $20$~KiB/token at 35B and $30$~KiB at
397B, and it \emph{grows with context} ($0.625$ and $0.94$~GiB/seq at a 32K context).
The GDN layers instead hold one fixed state, $n_{\text{GDN}} \times n_{\text{v-head}}
\times d_v \times d_k \times 2$~B (bf16) per sequence: $30$ and $90$~MiB/seq, forever,
which is why hybrids are attractive.

\textbf{Speculative verification: where GDN blows up.}
Speculation inverts the picture. To verify $k$ draft tokens (or a tree of $N$
nodes):
\begin{itemize}
\item \textbf{Softmax layers stay cheap.} Draft KV entries are appended, and
  rejected ones are dropped by moving a pointer, i.e., rollback is free. A tree even
  \emph{shares} its common-prefix KV across branches. The marginal cost is a
  handful of KiB.
\item \textbf{GDN layers explode during verification.} Verification is one parallel forward pass over all $k$ tokens at once. 
  In that single pass the GDN recurrent state advances all the way to the end (past token $k$) before any acceptance is checked.
  And so, we need to rollback to the state at the accepted node. The standard fix,
  which we call \emph{full-state snapshotting}, keeps the committed state plus one
  full snapshot per draft position: $k{+}1$ blocks per sequence per GDN layer for a
  chain of $k$ draft tokens (a tree of $N$ nodes needs $N{+}1$, one per node). For
  $k=3$ that is $4\times$ the committed state, $120$~MiB/seq (35B) and $360$~MiB/seq
  (397B). Moreover, the snapshots are \emph{unshareable across branches}. Snapshot memory 
  scales with draft tokens, caps concurrency and rules out wide trees. This is the bottleneck we remove.
\end{itemize}

\textbf{Existing approaches and trade-offs.}
\begin{itemize}
\setlength{\itemsep}{1pt}\setlength{\parskip}{0pt}
\item \textbf{(i) Full-state snapshotting.} The current default in \emph{both}
vLLM~\citep{kwon2023vllm} (\texttt{storeall} in our figures) and SGLang~\citep{zheng2024sglang}: one recurrent-state snapshot per draft
position, rolled back by indexing to the accepted one. Rollback is free; memory scales
with draft size and snapshots cannot be shared across tree branches.
\item \textbf{(ii) Deferred materialization with a rank-1 cache}
(ReplaySSM~\citep{dao2026replayssm}, shipping
concurrently July 2026 in a vLLM RFC and TensorRT-LLM) keeps a checkpoint plus a short
cached history and defers materializing the state until a periodic flush, so state is
written far less often than every draft step. Inside the verify window it solves the
\emph{same} triangular system we do (Section~\ref{sec:method}), so the distinction is
\emph{when} state is materialized, not how it is computed. It covers Mamba2 and GDN, and
the implementation linked in the RFC is chain-only (not a tree). It is the closest prior point
in vLLM; we benchmark against it in Appendix~\ref{app:replayssm}. ReplaySSM has also
been incorporated in SGLang~\citep{sglang_replayssm_issue} as an opt-in feature.
\item \textbf{(iii) Transition-matrix tree recompute}
(STree~\citep{wu2025stree,wu2024snakes}) verifies a tree, but only
for \emph{Mamba2}: its transition is a bare \emph{scalar} decay, so its speedup rests
on turning a product of gates into a cheap cumulative sum. This does
\emph{not} exist for GDN's non-commuting matrix transition (scalar decay $\times$
rank-1 correction).
\item \textbf{(iv) Tree-structured closed form in SGLang}
(Bole~\citep{wang2026bole}, concurrent, Aug.\ 2026)
is the most direct overlap with this paper: it also rewrites the linear-attention
recurrence into a tree-structured closed form so a whole draft tree can be verified in
parallel rather than branch by branch, reporting $82$--$99\times$ transient-memory
reduction and $3.4$--$7.7\times$ faster tree verification from its kernel. It targets
SGLang rather than vLLM and, by its own description, is framed at the level of
hybrid-attention recurrences generally. We have not run a head-to-head, since Bole's
paper does not link to a code implementation; it is the closest concurrent work and the
comparison we most want to make next.
\end{itemize}

Our method is the tree-structured WY/UT transform of the gated delta rule
(Section~\ref{sec:method}), implemented in vLLM (Section~\ref{sec:impl}): one triangular
solve verifies every draft node. It generalizes DeltaNet's~\citep{yang2024delta} chain WY transform to a tree
strict-ancestor form reused for verification and rollback. A single small pseudo-value
matrix written only on commit stands in for the per-node snapshots. (iv)
is architecturally closest to us and was arrived at independently on a different
serving stack.

\section{Method: TreeWY}
\label{sec:method}

\textbf{Key rewrite.} Expanding \eqref{eq:gdn} shows the delta rule is \emph{decayed
additive attention with a corrected value}: $S_t = \alpha_t S_{t-1} + \tilde v_t
k_t^\top$ with \emph{pseudo-value} $\tilde v_t = \beta_t(v_t - \alpha_t S_{t-1} k_t)$
(raw value minus what the state already predicts). Given the $\tilde v_t$, both the state
and every output are decay-weighted sums over the preceding tokens, so obtaining the
$\tilde v_t$ without walking the recurrence removes the need for intermediate states.

\textbf{Verifying a draft.} A draft can be a chain or a tree: a chain proposes one
candidate per position, a tree lets a node fan out into several. Either way, we lay out
its $N$ nodes in DFS pre-order from the committed state $S_0$, so every ancestor of node
$t$ has a smaller index ($i \prec t$; on a chain this is just $i<t$, and $N{=}k$).
Chaining the per-token rewrite along every root-to-node path turns verification into one
linear system for all the pseudo-values at once, $\tilde V \in \mathbb{R}^{N \times d_v}$:
\begin{equation}
\bigl(I + \operatorname{diag}(\beta)\,G\bigr)\,\tilde V = R,\quad
G[t,i] = \tfrac{g_t}{g_i}(k_t^\top k_i)\,[\,i \prec t\,],\quad
R[t] = \beta_t v_t - \beta_t g_t\,(S_0 k_t), \label{eq:solve}
\end{equation}
with $g_t$ the cumulative decay from $S_0$ to $t$. Parents precede children, so $G$ is
strictly lower-triangular, and the whole draft solves in one forward substitution: no
recurrence, no per-node state, and no change to the system as the tree widens. This is
DeltaNet's~\citep{yang2024delta} WY/UT transform applied to the whole verify window at once; every node's
output reads out from that same solve.

\textbf{Reconstruct on commit.} Once verification accepts node $a$, we rebuild the
continuation state directly from $\tilde V$, summing over $a$'s ancestors:
\begin{equation}
S_a = g_a S_0 + \sum_{i \preceq a} \tfrac{g_a}{g_i}\,\tilde v_i k_i^\top,
\label{eq:commit}
\end{equation}
which becomes the next round's $S_0$. We store only $\tilde V$, $O(N d_v)$, instead of
one full state per node: a $128\times$ smaller object per head at $d_k{=}d_v{=}128$,
which is what turns snapshotting's $N{+}1$ state blocks per sequence per layer into one,
for a chain or a tree alike. The scalar-decay ancestor mask is
STree's~\citep{wu2025stree}; extending it to
carry GDN's rank-1 delta correction, so it covers GDN's non-commuting transition rather
than only Mamba2's scalar one, is the piece we add.

\section{Implementation}
\label{sec:impl}

We implement TreeWY for vLLM, as a fork of its main branch (not yet upstreamed as of this writing) through two \texttt{SpeculativeConfig} options:
\texttt{mamba\_state\_commit="reconstruct"} turns on our commit strategy in place of the
default \texttt{"store\_all"}, and \texttt{draft\_tree\_widths} turns a chain into a tree
by giving the branching factor per level. All
evaluations here run with prefix caching disabled. A chain (or
an all-ones tree, which is really just a chain) verifies and commits in one fused,
CUDA-graph-capturable Triton kernel. A real tree ($w{>}1$) needs a non-causal ancestor
mask that cannot be replayed from a CUDA graph, so vLLM drops the whole model to
piecewise capture, which also evicts the GDN mixer from graphs, making the cost far
larger than the mask itself. Trees must also be scheduled \emph{atomically}: a DFS
prefix of a tree is a different topology, so a request whose tree does not fit the
per-step token budget skips speculation rather than being truncated. All results use
greedy drafting and verification. We check correctness two ways: the closed form matches the
per-node recurrence to $\sim\!10^{-15}$ (fp64) and $\sim\!10^{-7}$ (fp32), and the
production kernel matches that reference within bf16 tolerance. Token
streams are therefore not bit-identical to the baseline; we gate correctness against a
shared no-speculation reference and compare the \emph{acceptance length} directly between
\texttt{treewy} and \texttt{storeall}.

\section{Evaluation}
\label{sec:eval}

\textbf{Setup.} We serve Qwen3.5-35B-A3B (tensor-parallel degree 1, TP1) and
Qwen3.5-397B-A17B (TP8) in vLLM on
B200 GPUs ($178$~GiB HBM/device) with a depth-3 MTP draft chain, comparing
\texttt{treewy} (ours) against snapshotting (vLLM's default, \texttt{storeall}). We
sweep GPU memory utilization (\texttt{gmu}) $\in\{0.6,0.75,0.9\}$ over six workloads
(ShareGPT~\citep{sharegpt}, spec-bench~\citep{xia2024specbench},
BurstGPT~\citep{wang2024burstgpt}, and synthetic balanced-chat, generation-heavy,
summarize-heavy).\footnote{Licenses: Qwen3.5~\citep{qwen2025tech} and
vLLM~\citep{kwon2023vllm} are Apache-2.0, as is spec-bench; BurstGPT is CC-BY-4.0;
the ShareGPT mirror we use is tagged Apache-2.0 on Hugging Face. The synthetic
workloads use vLLM's own \texttt{random} benchmark generator.} Five of
the six sweep fixed max-concurrency
$\in\{1,8,32,64,128,256\}$; BurstGPT instead sweeps its own Poisson arrival-rate
$\in\{4,8,16,32,64\}$~req/s, a different independent variable.

\begin{table}[t]
\begin{minipage}{0.62\linewidth}
\centering
\caption{Concurrency sweep at each model's tightest measured budget (35B:
\texttt{gmu}~0.6; 397B: \texttt{gmu}~0.75), geomean over 5 workloads, $>\!1$ favours
TreeWY. \emph{tput} is output throughput; \emph{KV red.} is peak KV-cache usage
reduction; \emph{TTFT} is p99 TTFT; \emph{TPOT} is mean
time-per-output-token (TPOT). $b$ is the ratio of peak \emph{admitted}
requests, the mechanism behind the other columns past the knee (remaining budgets:
Table~\ref{tab:conc-rest}).}
\label{tab:conc}
\tiny
\setlength{\tabcolsep}{2pt}
\begin{tabular}{lccccccccccc}
  \toprule
  & \multicolumn{5}{c}{\emph{35B-A3B} (TP1, \texttt{gmu} 0.6)} & &
    \multicolumn{5}{c}{\emph{397B-A17B} (TP8, \texttt{gmu} 0.75)} \\
  \cmidrule(lr){2-6}\cmidrule(lr){8-12}
  conc. & tput & KV red. & TTFT & TPOT & $b$ & &
          tput & KV red. & TTFT & TPOT & $b$ \\
  \midrule
  1   & 0.94 & 2.43 & 1.02 & 0.94 & 1.00 & & 0.93 & 2.00 & 0.95 & 0.94 & 1.00 \\
  8   & 1.02 & 2.40 & 1.03 & 1.02 & 1.00 & & 0.97 & 2.19 & 1.08 & 0.96 & 1.00 \\
  32  & 1.01 & 2.45 & 1.02 & 1.00 & 1.00 & & 1.01 & 2.19 & 0.99 & 1.01 & 1.00 \\
  64  & 1.00 & 2.44 & 1.01 & 0.99 & 1.01 & & 1.01 & \textbf{2.19} & 1.00 & 1.02 & 1.00 \\
  128 & \textbf{1.20} & 1.64 & \textbf{5.62} & 0.83 & 1.45 & & 1.02 & 2.09 & 1.07 & 0.99 & 1.04 \\
  256 & \textbf{1.40} & 1.04 & 3.97 & 0.60 & \textbf{2.50} & & \textbf{1.15} & 1.61 & \textbf{3.35} & 0.90 & \textbf{1.24} \\
  \bottomrule
\end{tabular}
\end{minipage}%
\hfill
\begin{minipage}{0.35\linewidth}
\centering
\caption{Tree width vs.\ acceptance length. \emph{Serving} sweep: Qwen3.5-35B-A3B,
TP1, \texttt{gmu} 0.9, spec\_bench ($240$ prompts, $256$ output tokens), greedy, no
prefix caching. \emph{acc.} is vLLM's serving acceptance length
$1+(\text{accepted}/\text{drafts})$, which \emph{includes} the bonus token and is
bounded by $4$ at depth 3. $N$ is the number of draft nodes; snapshotting would hold $N{+}1$ state
blocks per layer per request at that shape, TreeWY holds $1$ at every row.}
\label{tab:widthacc}
\scriptsize
\setlength{\tabcolsep}{4pt}
\begin{tabular}{lrr}
  \toprule
  shape & $N$ & acc. \\
  \midrule
  \texttt{storeall} chain & 3  & 3.24 \\
  \texttt{treewy} chain & 3  & 3.23 \\
  \midrule
  \multicolumn{3}{l}{\texttt{treewy} tree} \\
  $(1,1,1)$   & 3  & 3.23 \\
  $(2,1,1)$   & 6  & 3.30 \\
  $(2,2,1)$   & 10 & 3.38 \\
  $(2,2,2)$   & 14 & 3.38 \\
  $(3,2,2)$   & 21 & 3.41 \\
  $(3,3,2)$   & 30 & 3.45 \\
  $(3,3,3)$   & 39 & 3.58 \\
  \bottomrule
\end{tabular}
\end{minipage}
\end{table}

\begin{figure}[h]
  \centering
  \includegraphics[width=\linewidth]{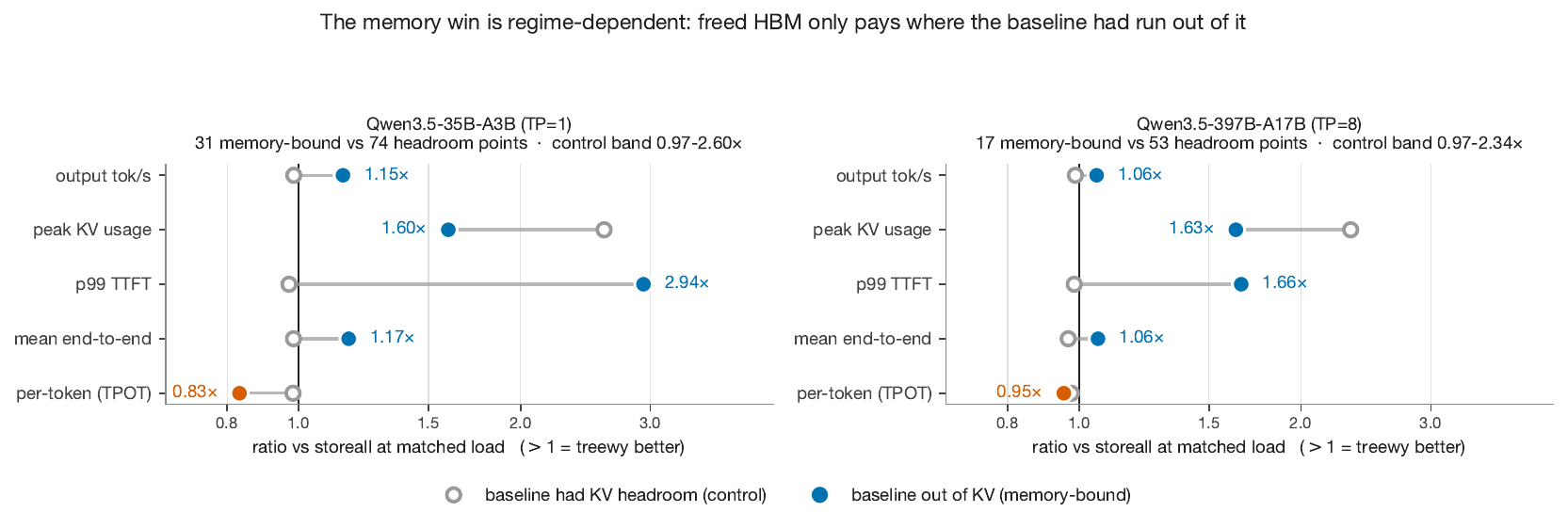}
  \caption{\texttt{treewy}/\texttt{storeall} ratios over all $175$ matched points (the
  concurrency sweeps of Tables~\ref{tab:conc} and~\ref{tab:conc-rest} plus BurstGPT), split
  by whether the \emph{baseline} had run out of KV headroom at that point, the same
  partition as Appendix~\ref{app:admission}. Open markers are the headroom control: every metric except
  KV usage sits within a few percent of parity there. Filled markers are where the
  freed HBM is actually spent: at 35B (31/105 points memory-bound) TreeWY reaches
  $1.15\times$ throughput, $2.94\times$ lower p99 TTFT and $1.17\times$ lower mean
  end-to-end latency at $1.60\times$ more KV headroom, for a $0.83\times$ per-token
  cost; 397B ($17/70$ points memory-bound) shows the same shape at smaller magnitude
  ($1.06\times$ throughput, $1.66\times$ lower p99 TTFT, $1.63\times$ more KV headroom,
  $0.95\times$ per-token cost), since its sweep reaches fewer points past the knee. This is the single figure that
  makes the paper's central qualifier legible at a glance: the win is not a fixed
  multiplier, it is conditional on memory pressure, and averaging over both regimes
  (as a single headline number would) hides exactly that.}
  \label{fig:regime}
\end{figure}

\textbf{Correctness.} A \emph{matched point} is one (dataset, concurrency, gmu) combination run under both \texttt{treewy} and \texttt{storeall}: $5$ workloads $\times$ $6$
concurrencies $+$ BurstGPT $\times$ $5$ arrival rates $= 35$ per \texttt{gmu}, times
$3$ \texttt{gmu} at 35B and $2$ at 397B gives $175$ total ($105$ at 35B, $70$ at
397B), the same partition Fig.~\ref{fig:regime} splits by memory pressure. Across
all of them, acceptance length is essentially identical to the baseline (mean
$|\Delta|=0.039$, max $0.33$), and per-position acceptance, averaged over all points,
matches to within $0.01$ at every depth.

\begin{figure}[h]
  \centering
  \includegraphics[width=\linewidth]{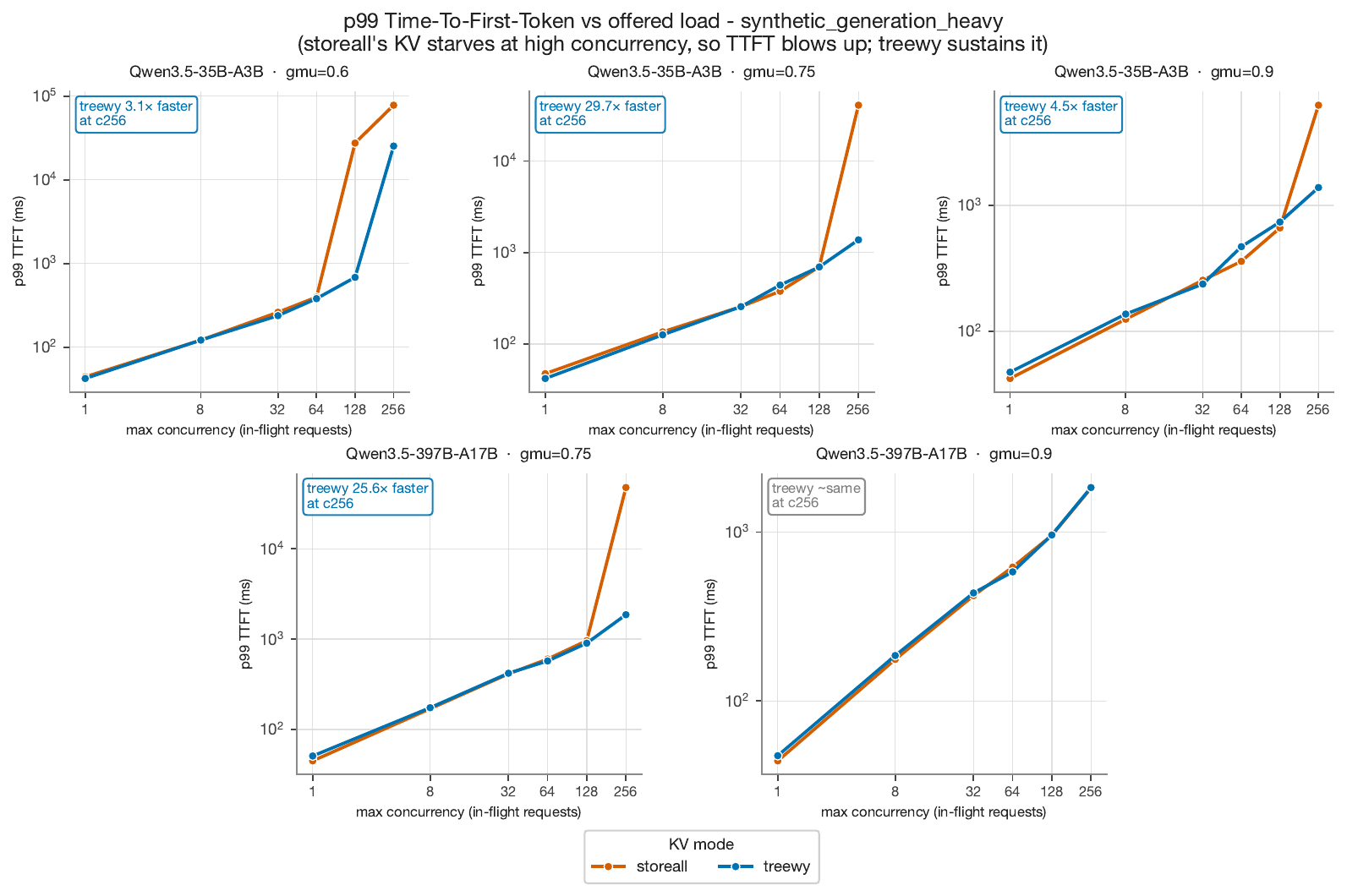}
  \caption{p99 time-to-first-token (TTFT)  vs.\ offered load (generation-heavy workload).
  As concurrency rises, the snapshotting baseline's KV pool saturates and it must
  queue requests, so its TTFT blows up (log scale): at 256 concurrency it is up to
  $\sim\!30\times$ (35B, utilization 0.75) and $26\times$ (397B, utilization 0.75) worse than
  TreeWY. The largest gap over the whole sweep, $40\times$ ($683$ vs.\ $27489$~ms), is
  at 128 concurrency (35B, utilization 0.6), where the baseline saturates but TreeWY
  still does not. The gap closes only when memory is slack (397B, utilization 0.9).}
  \label{fig:ttft}
\end{figure}

\textbf{Memory and throughput.} TreeWY's peak KV usage is $2$--$3\times$ lower at the
same load (one state block per layer vs.\ $k{+}1$) until it spends that headroom on
admitting more requests ($b>1$), so it preempts less under
pressure ($1314$ vs.\ $1622$ preempted requests at 35B, $51$ vs.\ $909$ at
397B). That headroom
drives Table~\ref{tab:conc} and Fig.~\ref{fig:ttft}: TreeWY wins outright where memory binds (up to
$1.49\times$ throughput and $\sim\!40\times$ lower p99 TTFT at a single point), and trails by a few percent
on throughput where it doesn't (geomean $0.985\times$ and $0.989\times$ over the $74$ and
$53$ points at 35B and 397B where the baseline never exceeded $0.9$ KV usage or queued),
still at $2$--$3\times$ less KV. That per-token dip is mostly the price of admitted concurrency, not a slower kernel (Appendix~\ref{app:admission}).
\\ The knee migrates predictably with the budget: 128 concurrency at
$\texttt{gmu}{=}0.6$, 256 at $0.75$, never at $0.9$ (Appendix~\ref{app:budgets}); and
wherever $b{=}1.00$ the freed memory is real but \emph{unspent}: KV reduction still holds at
$2.0$--$2.5\times$ while throughput and TPOT sit within $8\%$ of parity and p99 TTFT
ranges $0.65$--$1.09\times$.
Fig.~\ref{fig:regime} decomposes every metric by memory pressure.
\\ We separately benchmark against ReplaySSM on the identical 35B sweep
(Appendix~\ref{app:replayssm}).

\textbf{Tree width.} Table~\ref{tab:widthacc} sweeps tree width $w$ at depth 3 against
a \texttt{storeall} chain baseline.
TreeWY holds one state block per layer regardless of $w$ (the same freed-memory
mechanism as Table~\ref{tab:conc}, now spent on width instead of concurrency), and
acceptance length keeps rising with $w$.
Width is therefore \emph{affordable} under reconstruction
where snapshotting can't afford it at all: a \texttt{storeall} baseline's per-request cost
grows with $N$ while TreeWY's stays flat at one block. It is not yet a throughput win,
though: a wider tree pushes $N{+}1$ tokens per step through the target and runs
piecewise rather than fused (Section~\ref{sec:impl}), so we report trees as enabled and
correct, not as a speedup.

\section{Conclusion}
Treating the gated delta rule as a tree-structured WY transform lets us verify a GDN draft
tree with one triangular solve and reconstruct only the accepted state on commit, trading
$O(N)$ state snapshots for one small pseudo-value matrix, on chains and trees alike. On
chains this is already a fused, graph-capturable kernel that turns freed HBM into higher
throughput and much lower TTFT wherever memory binds, at a residual per-step cost that is
an implementation artifact rather than a property of the closed form. On trees the same
mechanism holds the stored state flat at one block regardless of width
(Table~\ref{tab:widthacc}), which makes a wider, higher-acceptance draft affordable,
though not yet a throughput win, since acceptance is capped by draft
depth and the non-capturable tree verify kernel costs more than the extra acceptance returns. Fusing the tree path into a single graph-capturable kernel and testing the
deferred-state-write levers that ReplaySSM and Bole both point to are the next immediate
engineering steps, alongside extending to a second model family.

\appendix

\clearpage
\section{Remaining memory budgets}
\label{app:budgets}

\begin{table}[h]
  \caption{The higher memory budgets, same axes and orientation as
  Table~\ref{tab:conc}. At 397B the knee appears only at $0.75$/256. Wherever
  $b$ reads $1.00$ the freed memory is real but unspent, as discussed in
  Section~\ref{sec:eval}.}
  \label{tab:conc-rest}
  \centering
  \small
  \setlength{\tabcolsep}{5pt}
  \begin{tabular}{lccccc}
    \toprule
    conc. & tput & KV red. & TTFT & TPOT & $b$ \\
    \midrule
    \multicolumn{6}{l}{\emph{Qwen3.5-35B-A3B}, \texttt{gmu} 0.75} \\
    \quad 1   & 0.95 & 2.32 & 1.09 & 0.94 & 1.00 \\
    \quad 8   & 0.98 & 2.44 & 1.00 & 1.00 & 1.00 \\
    \quad 32  & 1.00 & 2.45 & 1.01 & 0.99 & 1.00 \\
    \quad 64  & 1.00 & \textbf{2.48} & 0.94 & 1.00 & 1.00 \\
    \quad 128 & 1.02 & 2.43 & 0.99 & 1.02 & 1.01 \\
    \quad 256 & \textbf{1.18} & 1.62 & \textbf{4.61} & 0.80 & \textbf{1.46} \\
    \midrule
    \multicolumn{6}{l}{\emph{Qwen3.5-35B-A3B}, \texttt{gmu} 0.9 (no knee)} \\
    \quad 1   & 0.93 & 2.32 & 0.91 & 0.93 & 1.00 \\
    \quad 8   & 0.97 & 2.44 & 0.65 & 0.98 & 1.00 \\
    \quad 32  & 0.98 & 2.45 & 1.06 & 0.98 & 1.00 \\
    \quad 64  & 0.98 & 2.46 & 0.94 & 0.98 & 1.00 \\
    \quad 128 & 1.00 & \textbf{2.48} & 0.99 & 0.99 & 1.00 \\
    \quad 256 & 1.03 & 2.26 & 1.44 & 0.95 & 1.07 \\
    \midrule
    \multicolumn{6}{l}{\emph{Qwen3.5-397B-A17B}, \texttt{gmu} 0.9 (no knee)} \\
    \quad 1   & 0.93 & 2.00 & 0.95 & 0.92 & 1.00 \\
    \quad 8   & 0.97 & 2.19 & 0.94 & 0.96 & 1.00 \\
    \quad 32  & 0.99 & 2.19 & 0.94 & 0.99 & 1.00 \\
    \quad 64  & 0.99 & 2.19 & 1.00 & 0.99 & 1.00 \\
    \quad 128 & 0.99 & \textbf{2.22} & 0.98 & 0.99 & 1.00 \\
    \quad 256 & 1.00 & 2.15 & 1.02 & 0.98 & 1.03 \\
    \bottomrule
  \end{tabular}
\end{table}
\FloatBarrier

\section{Admission vs.\ per-step cost}
\label{app:admission}

\textbf{Reading the latency columns.} Admitting more requests raises \emph{per-request}
latency at unchanged per-step cost, so a TPOT ratio below $1$ is partly the price of that
concurrency: across the 35B points the penalty tracks batch growth almost exactly (Pearson
$r=-0.88$ against the running-batch ratio, $+0.88$ for throughput). At the 31 points where
the baseline saturated, TreeWY pays $0.83\times$ on TPOT yet its mean \emph{end-to-end}
latency still improves ($1.17\times$), because requests the baseline could not admit stop
waiting; at the other 74 its batch is matched to within $1\%$ and TPOT is $0.98$; that,
not the aggregate, is our per-step cost (same decomposition for ReplaySSM, $r=-0.79$).

\section{ReplaySSM comparison}
\label{app:replayssm}

\textbf{Comparison with ReplaySSM.} ReplaySSM~\citep{dao2026replayssm} (approach (ii)) is the closest concurrent
point with a chain-mode implementation, so we ran the identical 35B sweep
against it (Tables~\ref{tab:decomp},~\ref{tab:conccmp} below). The two arms sit on
different vLLM builds, so each is normalised to the snapshotting baseline from
\emph{its own} image; Table~\ref{tab:decomp} lists both baselines so the drift is
visible: across all $105$ points the \texttt{storeall} throughputs agree to within $1\%$
(geomean of per-point ratios $0.993$; $2.6\%$ on the 256-concurrency slice of
Table~\ref{tab:decomp}), so the two arms'
own-baseline ratios below are directly comparable to each other. Both delete the
per-position snapshots and
free the same order of KV headroom, TreeWY marginally more ($0.96\times$ ReplaySSM's
peak usage), and both preserve acceptance on real prompts (mean $|\Delta|$ acceptance
length $0.029$ vs.\ $0.037$). ReplaySSM turns its headroom into more
throughput on a \emph{chain} ($1.12$--$1.20\times$ its baseline vs.\ our
$0.99$--$1.08\times$; geomean per memory budget) and preempts less often. That margin is not the solve (the
solve is shared): our working hypothesis is that it is the scheduling around it, since
they defer writing state on a verify step where we materialize the accepted state on
every commit, but this is an attribution, not a validated claim: we have not
prototyped a deferred-write commit path ourselves to check that it actually closes the
gap. If it holds, it is an implementation lever open to us and orthogonal to the closed
form; testing it, alongside the analogous sweep we have not yet run against
Bole~\citep{wang2026bole} (approach (iv), Section~\ref{sec:memory}, which targets a
different serving engine), is future work.

\begin{table}[h]
  \caption{Why the ratios differ, decomposed at 256 concurrency (Qwen3.5-35B-A3B, all 15
  points: 5 workloads (ShareGPT, spec-bench, synthetic balanced-chat,
  generation-heavy, summarize-heavy) $\times$ 3 \texttt{gmu}; BurstGPT excluded, its
  own independent variable). Geomeans of the \emph{raw} values over the $15$ points
  first, then each method's ratios to its own baseline and the batch $\div$ cost
  check. Both
  methods admit essentially the same work: 238 vs 224 peak running requests, and at
  \texttt{gmu}~0.6 per workload they are indistinguishable (253/255, 254/255,
  256/256, 256/256, 109/107 for ShareGPT/spec-bench/balanced-chat/generation-heavy/
  summarize-heavy), because both simply reach the offered concurrency once the
  snapshots are gone. TreeWY's larger
  \emph{admitted batch} ratio below ($1.577$ vs.\ $1.458$) is therefore a denominator artifact: its
  own baseline happened to admit slightly fewer at \texttt{gmu}~0.6 (92 vs 97 on balanced-chat, 61 vs 63 on
  summarize-heavy). What separates the two is per-token cost, and once both are saturated
  and decode-bound the throughput ratio is just $\text{batch} \div \text{cost}$, which
  closes to within $3\%$ for both. So at this operating point the memory mechanism is a
  tie and the whole difference is per-step efficiency.}
  \label{tab:decomp}
  \centering
  \small
  \setlength{\tabcolsep}{5pt}
  \begin{tabular}{lrrrr}
    \toprule
    & \multicolumn{2}{c}{Ours} & \multicolumn{2}{c}{ReplaySSM} \\
    \cmidrule(lr){2-3}\cmidrule(lr){4-5}
    & \texttt{storeall} & TreeWY & \texttt{storeall} & ReplaySSM \\
    \midrule
    peak running requests   & 151.1 & 238.2 & 153.6 & 223.9 \\
    output tok/s            & 8469  & 10094 & 8250  & 11357 \\
    mean TPOT (ms)          & 16.4  & 21.2  & 17.0  & 17.6 \\
    p99 TTFT (ms)           & 8167  & 2746  & 8965  & 2958 \\
    \midrule
    \multicolumn{5}{l}{\emph{ratio vs own baseline, and the batch $\div$ cost check}} \\
    admitted batch          & \multicolumn{2}{c}{$1.577$} & \multicolumn{2}{c}{$1.458$} \\
    per-token cost          & \multicolumn{2}{c}{$1.295$} & \multicolumn{2}{c}{$1.035$} \\
    batch $\div$ cost (predicted) & \multicolumn{2}{c}{$1.218$} & \multicolumn{2}{c}{$1.408$} \\
    throughput (measured)   & \multicolumn{2}{c}{$\mathbf{1.192}$} & \multicolumn{2}{c}{$\mathbf{1.376}$} \\
    \bottomrule
  \end{tabular}
\end{table}

\begin{table}[h]
  \caption{The same load axis for both methods (Qwen3.5-35B-A3B), each normalised to
  the \texttt{storeall} baseline from \emph{its own} vLLM build, so the two columns are
  comparable even though the builds are not. Each row is a geomean over the $15$ points
  ($5$ workloads $\times$ $3$ \texttt{gmu}) at that concurrency. \emph{Per-token} speedup is based on mean
  TPOT. ReplaySSM leads on throughput at every
  load, and the gap is already present where neither method is admitting anything extra
  (batch $1.00$, loads 1--64), which is what identifies it as per-step cost rather than
  a memory effect. Past the knee the two converge on the memory-driven columns: at 256
  concurrency the peak-KV reduction is nearly equal ($1.56$ vs.\ $1.57$), i.e.\ both
  free the same headroom, and Table~\ref{tab:decomp} shows they spend it the same way
  ($238$ vs.\ $224$ peak running requests). On the BurstGPT arrival-rate sweep at $64$~req/s the p99 TTFT
  improvements are $10.72\times$ and $10.20\times$ respectively. The per-token column is the
  clearest read on per-step cost: ReplaySSM leads it at every load, and
  Table~\ref{tab:decomp} decomposes the 256-concurrency point.}
  \label{tab:conccmp}
  \centering
  \small
  \setlength{\tabcolsep}{6pt}
  \begin{tabular}{lcccc}
    \toprule
    & Output tput. & Peak KV & p99 TTFT & Per-token \\
    concurrency & gain $\uparrow$ & reduction $\uparrow$ & speedup $\uparrow$ & speedup $\uparrow$ \\
    & ours / Replay & ours / Replay & ours / Replay & ours / Replay \\
    \midrule
    1   & 0.94 / \textbf{1.06} & \textbf{2.36} / 2.15 & \textbf{1.00} / 0.98 & 0.94 / \textbf{1.07} \\
    8   & 0.99 / \textbf{1.08} & \textbf{2.43} / 2.26 & 0.87 / \textbf{1.20} & 1.00 / \textbf{1.07} \\
    32  & 1.00 / \textbf{1.11} & \textbf{2.45} / 2.29 & \textbf{1.03} / 0.99 & 0.99 / \textbf{1.11} \\
    64  & 0.99 / \textbf{1.17} & \textbf{2.46} / 2.29 & 0.96 / \textbf{1.01} & 0.99 / \textbf{1.17} \\
    128 & 1.07 / \textbf{1.27} & \textbf{2.15} / 2.02 & \textbf{1.76} / 1.60 & 0.94 / \textbf{1.14} \\
    256 & 1.19 / \textbf{1.38} & 1.56 / \textbf{1.57} & 2.97 / \textbf{3.03} & 0.77 / \textbf{0.97} \\
    \bottomrule
  \end{tabular}
\end{table}

% Flush Tables 4-5 and start this appendix on a fresh page.
\clearpage
\section{Additional serving curves}

\begin{figure}[h]
  \centering
  \includegraphics[width=\linewidth]{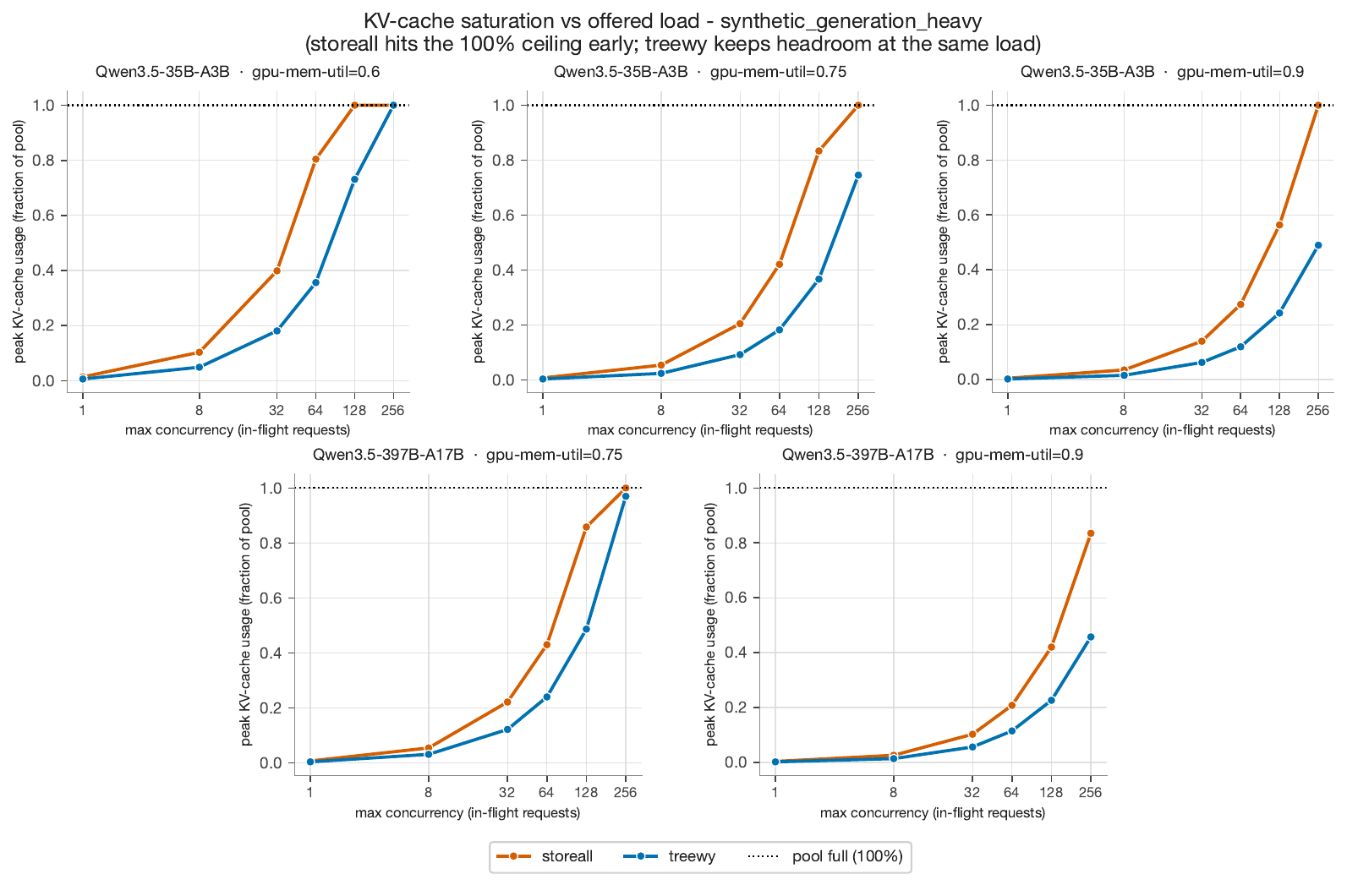}
  \caption{Peak KV-cache usage (fraction of pool) vs.\ offered load (generation-heavy workload), per model
  and GPU memory utilization (\texttt{gmu}). The snapshotting baseline (\texttt{storeall}) hits
  the 100\% ceiling early and must then queue the requests it cannot fit; TreeWY uses
  about $2\times$ less of the KV pool at the same load, until it admits more requests. That freed capacity buys the admission,
  throughput and TTFT wins (Fig.~\ref{fig:ttft}).}
  \label{fig:kv}
\end{figure}

\FloatBarrier
\section{Compute resources}
\label{app:compute}

\textbf{GPU-hours.} All runs use B200 GPUs ($178$~GiB HBM/device). Each job serves one
(model, mode, \texttt{gmu}) combination through its full $35$-point concurrency/arrival-rate
sweep in one continuous server session. The 35B main
sweep (Table~\ref{tab:conc}, Appendix~\ref{app:budgets}) is $6$ jobs
(\texttt{storeall}/\texttt{treewy} $\times$ $3$ \texttt{gmu}) at TP1, $\approx\!11.8$
GPU-hours; the 397B main sweep is $4$ jobs (\texttt{storeall}/\texttt{treewy} $\times$ $2$
\texttt{gmu}) at TP8, $\approx\!62.5$ GPU-hours; the ReplaySSM comparison
(Appendix~\ref{app:replayssm}) is $6$ further 35B/TP1 jobs, $\approx\!11.1$ GPU-hours.
Total: $\approx\!85$ GPU-hours across the three reported sweeps. This is a lower bound: it
excludes one-time model-loading/server-startup overhead at the start of each sweep
($\sim\!1$h for the 397B model, $\sim\!30$min for 35B), which does not recur across jobs
within the same sweep.

\end{document}